\PassOptionsToPackage{table,x11names}{xcolor}
\documentclass{article}

\usepackage{bookmark}
\PassOptionsToPackage{numbers, compress}{natbib}

    \usepackage[preprint]{neurips_2025}

\usepackage{tcolorbox}

\usepackage[utf8]{inputenc} 
\usepackage[T1]{fontenc}    
\usepackage{hyperref}       
\usepackage{url}            
\usepackage{booktabs}       
\usepackage{amsfonts}       
\usepackage{nicefrac}       
\usepackage{microtype}      
\usepackage{xcolor}       

\usepackage{graphicx}
\usepackage{subcaption} 
\usepackage{float}      

\hypersetup{
colorlinks= true,
urlcolor = blue,
linkcolor = red,
citecolor = blue,
}

\usepackage{xspace}
\usepackage{graphicx}
\usepackage{multicol}
\usepackage{multirow}
\usepackage{makecell}
\PassOptionsToPackage{numbers}{natbib}
\usepackage{wrapfig}
\usepackage{subcaption}
\usepackage{enumitem}
\usepackage{titletoc}
\usepackage{amsmath}
\usepackage{amssymb}
\usepackage{booktabs}    
\usepackage{longtable}   
\usepackage{array}       
\usepackage{pifont}

\usepackage{listings}
\setlist[itemize]{leftmargin=*}

\title{Performance of the SafeTerm AI-Based MedDRA Query System Against Standardised MedDRA Queries}

\author{%
\textbf{Francois Vandenhende$^{1}$ \quad Anna Georgiou$^{1}$ \quad Michalis Georgiou$^{1}$}\\
\textbf{Theodoros Psaras$^{1}$ \quad Ellie Karekla$^{1}$ \quad Elena Hadjicosta$^{1}$} \\[6pt]
$^1$ClinBAY Limited, Limassol, Cyprus\\
Correspondence: \texttt{francois@clinbay.com}\\
\url{https://app.clinbay.com/safeterm}
}

\PassOptionsToPackage{numbers}{natbib}

\begin{document}

\maketitle

\begin{abstract}
\noindent \textbf{Background:} In pre-market drug safety review, grouping related adverse event terms into Standardised MedDRA Queries (SMQ) or the FDA Office of New Drugs Custom Medical Queries (OCMQs) is critical for signal detection.\\
\textbf{Objective:} We assess the performance of SafeTerm Automated Medical Query (AMQ) on MedDRA SMQs. The AMQ is a novel quantitative artificial intelligence system that understands and processes medical terminology and automatically retrieves relevant MedDRA Preferred Terms (PTs) for a given input query, ranking them by a relevance score (0--1) using multi-criteria statistical methods.\\
\textbf{Methods:} The system (SafeTerm) embeds medical query terms and MedDRA PTs in a multidimensional vector space, then applies cosine similarity, and extreme-value clustering to generate a ranked list of PTs. Validation was conducted against tier-1 SMQs (110 queries, v28.1). Precision, recall and F1 were computed at multiple similarity-thresholds, defined either manually or using an automated method.\\
\textbf{Results:} High recall (\(\sim94\%\)) is achieved at moderate similarity thresholds, indicative of good retrieval sensitivity. Higher thresholds filter out more terms, resulting in improved precision (up to 89\%). The optimal threshold (\(\sim0.70\)) yielded an overall recall of \(\sim48\%\) and precision of \(\sim45\%\) across all 110 queries. Restricting to narrow-term PTs achieved slightly better performance at an increased (+0.05) similarity threshold, confirming increased relatedness of narrow versus broad terms. The automatic threshold (0.66) selection prioritizes recall (0.58) to precision (0.29).\\
\textbf{Conclusions:} SafeTerm AMQ achieves comparable, satisfactory performance on SMQs and sanitized OCMQs. It is therefore a viable supplementary method for automated MedDRA query generation, balancing recall and precision. We recommend using suitable MedDRA PT terminology in query formulation and applying the automated threshold method to optimise recall. Increasing similarity scores allows refined, narrow terms selection.
\end{abstract}


\section{Introduction}
The Medical Dictionary for Regulatory Activities\textsuperscript{\textregistered} (MedDRA\textsuperscript{\textregistered}) is the internationally recognised terminology system for coding adverse event (AE) terms in clinical trials and pharmacovigilance \cite{brown1999meddra}. A crucial safety signal-detection task is grouping semantically related AE terms (which may appear under different preferred terms or hierarchies) so that meaningful adverse-event clusters can be analysed consistently (e.g., “Hypoglycaemia” vs. “Blood glucose decreased”). Failure to group related events can dilute the apparent incidence of safety signals.

Traditionally, safety experts use Standardised MedDRA Queries (SMQs) \cite{mozzicato2007smq} to define clinically meaningful sets of MedDRA terms for signal detection and evaluation. Each SMQ captures a medical concept of interest, such as Hepatic disorders, by grouping relevant Preferred Terms (PTs) based on clinical, pathophysiological, or pharmacological criteria. They include broad and narrow term groupings to balance sensitivity and specificity. Some SMQs are organized hierarchically, where Tier 1 SMQs represent broad medical concepts, while Tier 2 to 5 SMQs define more specific sub-queries. As of MedDRA v28.1, there are 110 Tier 1 SMQs in production, covering a wide range of adverse-event domains.

Building and maintaining clinically meaningful query sets is a resource-intensive process, particularly as MedDRA continues to evolve. Recent advances in natural language processing (NLP), notably large language models (LLMs) and embedding-based representations, enable more scalable and consistent ways to generate or update such groupings. Previous studies have demonstrated that embedding models can enhance MedDRA term retrieval \cite{siegersma2022pipeline} and that cosine-similarity searches within embedding spaces can aid adverse event normalization \cite{lahiri2025benchmarking}.

In a previous study \cite{vandenhende2025ocmq}, we introduced SafeTerm, an AI-assisted pipeline for automated MedDRA term mapping, and demonstrated its performance against the FDA OCMQ \cite{proestel2025ocmq, fda2025ocmq} reference standard, achieving balanced performance comparable to expert-defined queries. In the present study, we extend this evaluation to MedDRA Standardised MedDRA Queries (SMQs) to assess the system’s usability and generalizability across standardized, hierarchical query structures used in pharmacovigilance practice.

\section{Materials and Methods}

\subsection{System Overview --- SafeTerm}
We developed the SafeTerm Medical Map, an AI-assisted framework that embeds PTs and generic drug names into a unified multidimensional vector space, with an interactive 2-D projection forming an intuitive “medical map.” SafeTerm supports a range of pre- and post-market pharmacovigilance applications, including MedDRA/ATC term coding, signal detection, query generation, and incidence pattern analysis using aggregated data from ClinicalTrials.gov. An online version is available to registered MedDRA users at \url{https://app.clinbay.com/safeterm}.

\subsection{Automated MedDRA Query (AMQ) Pipeline}
This component of SafeTerm automatically retrieves and ranks PTs relevant to any input medical query. The AMQ algorithm operates deterministically and is fully reproducible. It was developed independently of any OCMQ or SMQ reference materials and was not fine-tuned using their gold-standard term sets. The overall workflow consists of three main stages:

\noindent\textbf{Query Matching and Embedding.} The input query is first embedded and projected onto the SafeTerm MedDRA Map to identify the most relevant MedDRA PTs. Initial retrieval employs fuzzy string matching with a high similarity cutoff. If a direct textual match exceeds this threshold, the best-scoring PT is selected. When no suitable lexical match is found, semantic similarity is computed via cosine similarity in the embedding space. Up to three top PTs are returned, and if multiple candidates are identified, their embeddings are averaged to construct a composite query representation.

\noindent\textbf{Similarity Scoring, Thresholding, and Clustering.} Pairwise cosine similarities are calculated between the query and the best PT embeddings, and all MedDRA PT embeddings. The resulting similarity distribution is analysed using a two-means clustering algorithm to partition terms into highly relevant and non-relevant groups. An additional step applies the Knee method \cite{kneedle} to automatically determine the optimal sim(best PT) threshold at which adding more terms no longer yields meaningful improvement. This data-driven threshold selection replaces manually fixed cutoffs, improving robustness across different query types and MedDRA versions.

\noindent\textbf{Term Ranking.} Finally, PTs retained above the selected threshold are rank-ordered by their sim(best PT) scores, producing a prioritized list of MedDRA PTs most semantically aligned with the input concept.

\subsection{Validation against SMQ}
We validated the AMQ pipeline against the SMQ v28.1 Tier-1 set of 110 medical-query concepts. For each input concept, the AMQ generated a list of candidate PTs, with similarity scores indicating term relevance. We then computed:

\begin{itemize}
  \item \textbf{True Positives (TP)}: PTs present in both SMQ gold list and AMQ list.
  \item \textbf{Precision} \(=\) TP / (number of PTs in AMQ list).
  \item \textbf{Recall} \(=\) TP / (number of PTs in the SMQ gold list).
  \item \textbf{F1} = harmonic mean of precision and recall.
\end{itemize}

These metrics were computed at multiple similarity cut-offs from 0.50 to 0.90, as well as at the limit defined using the automated similarity cut-off method. We also analysed performance in the subgroup of Narrow PTs only. Summary statistics (mean/SD/min/max) were reported across all 110 SMQs.

\section{Results}

\subsection{AMQ Retrieval Performance}
\begin{figure}[H]
  \centering
  \includegraphics[width=0.9\textwidth]{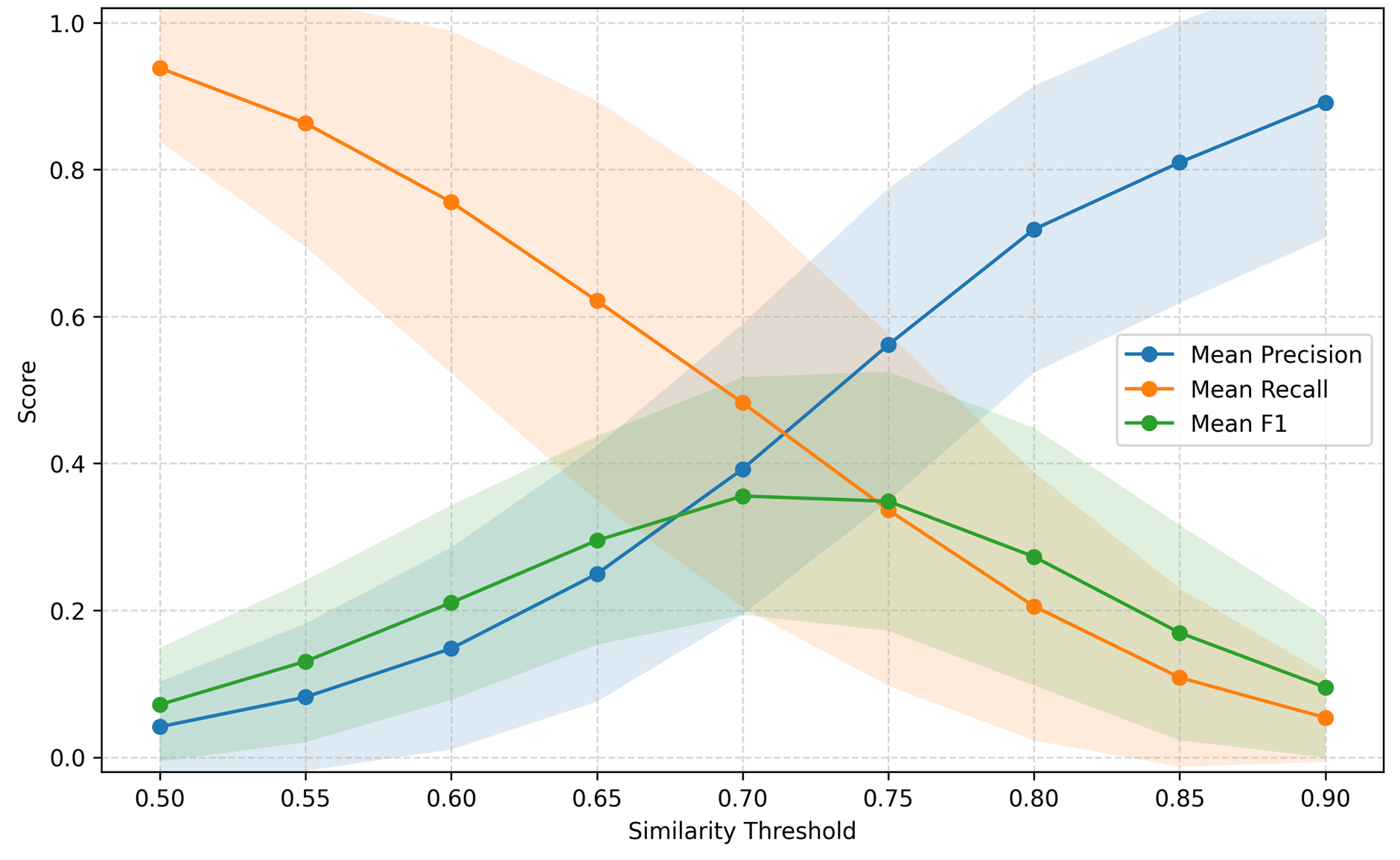}
  \caption{Mean $\pm$ SD: Precision, Recall, F1 versus similarity cut-off (across 110 Tier-1 SMQs).}
  \label{fig:perf_vs_cutoff}
\end{figure}

Summary results for retrieval performance (precision, recall, F1) across the 110 SMQs as a function of similarity cut-off are reported below, in Figure~\ref{fig:perf_vs_cutoff} and Table~\ref{tab:performance_summary}.

Key observations include:
\begin{itemize}
  \item At a similarity cut-off of 0.50, mean recall was approximately 0.94, indicating high sensitivity of the AMQ retriever.
  \item The optimal F1 score (\(\sim0.36\), SD = 0.16) occurred around a cut-off of 0.70, corresponding to a mean precision of 0.39 (SD = 0.20) and mean recall of 0.48 (SD = 0.28).
  \item At the highest threshold (0.90), mean precision reached 0.89 (SD = 0.18).
\end{itemize}

As expected, precision increased and recall decreased as the similarity threshold became more stringent. The balance between sensitivity and specificity, as indicated by the F1 score, was maximized at a threshold of approximately 0.70.

Table~\ref{tab:automated_vs_f1} compares the automated similarity threshold selection (Knee method) with the manually determined threshold yielding maximum F1. The automated approach recommends a slightly lower similarity cut-off, emphasising higher recall (+10\%) at the expense of lower precision (-16\%). This trade-off favours broader term retrieval, which may be desirable in exploratory signal detection contexts.
\subsection{Correlation with SMQ size and special cases}
\begin{figure}[H]
\centering
\begin{subfigure}[b]{0.48\textwidth}
    \centering
    \includegraphics[width=1\textwidth]{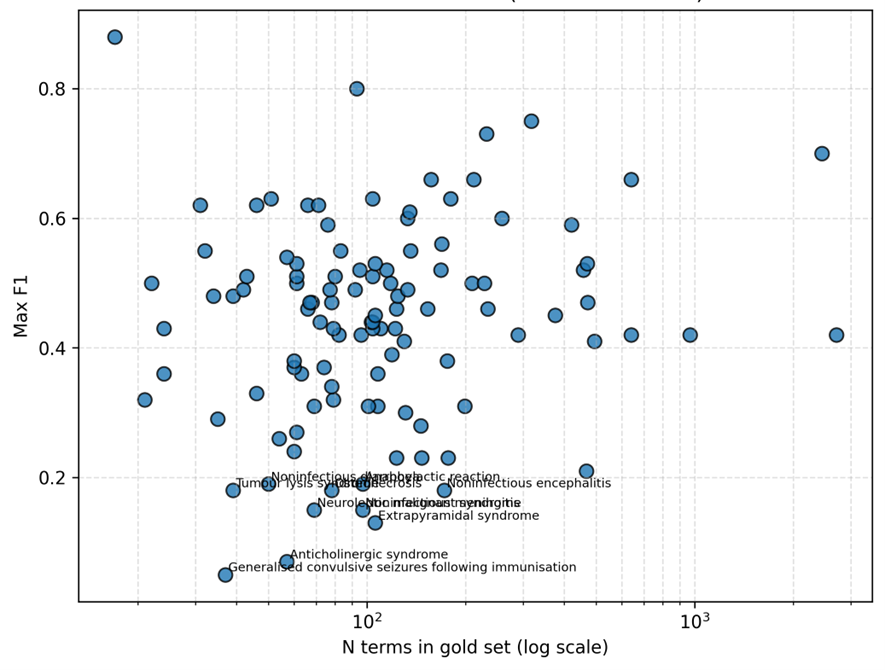}
    \caption{Narrow + Broad}
    \label{fig:fig02a}
\end{subfigure}
\hfill
\begin{subfigure}[b]{0.48\textwidth}
    \centering
    \includegraphics[width=1.07\textwidth]{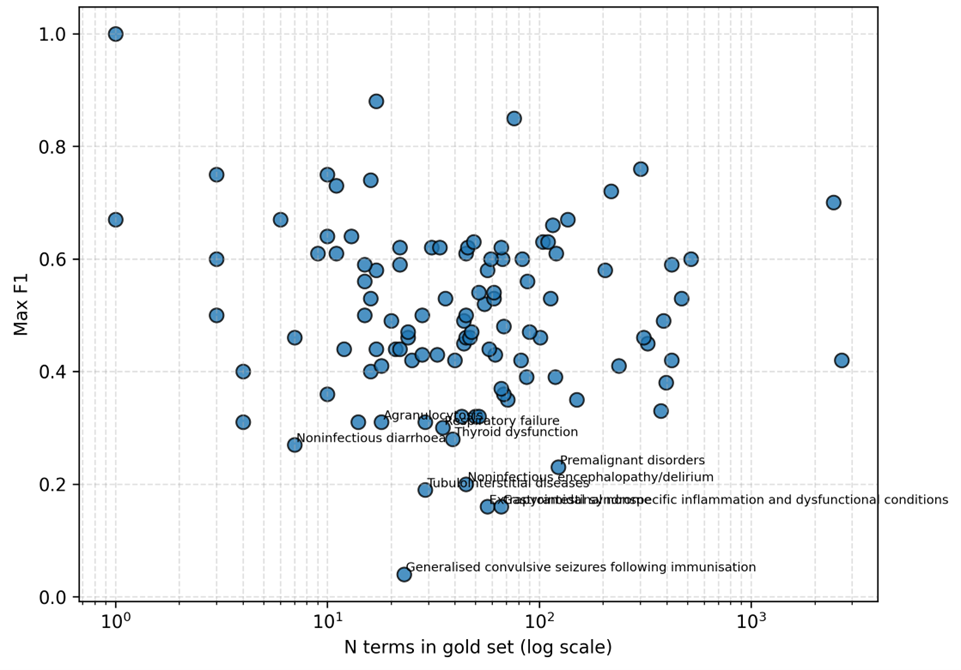}
    \caption{Narrow only}
    \label{fig:fig02b}
\end{subfigure}
\caption{Maximum F1 score versus SMQ gold set size for Narrow + Broad and Narrow-only Term Lists.}
\label{fig:narrow_broad}
\end{figure}

As shown in Figure~\ref{fig:narrow_broad}, the maximum F1 score showed little correlation with the number of terms in the SMQ reference sets (Pearson \(r=0.14\) and \(0.05\) for two analyses), suggesting that AMQ performance was consistent across both small and large query lists. Two low-performing SMQs were attributable to the presence of input concepts that were not valid MedDRA PTs; substituting the correct PTs markedly improved performance.

\subsection{Narrow-term retrieval}
\begin{figure}[H]
  \centering
  \includegraphics[width=0.9\textwidth]{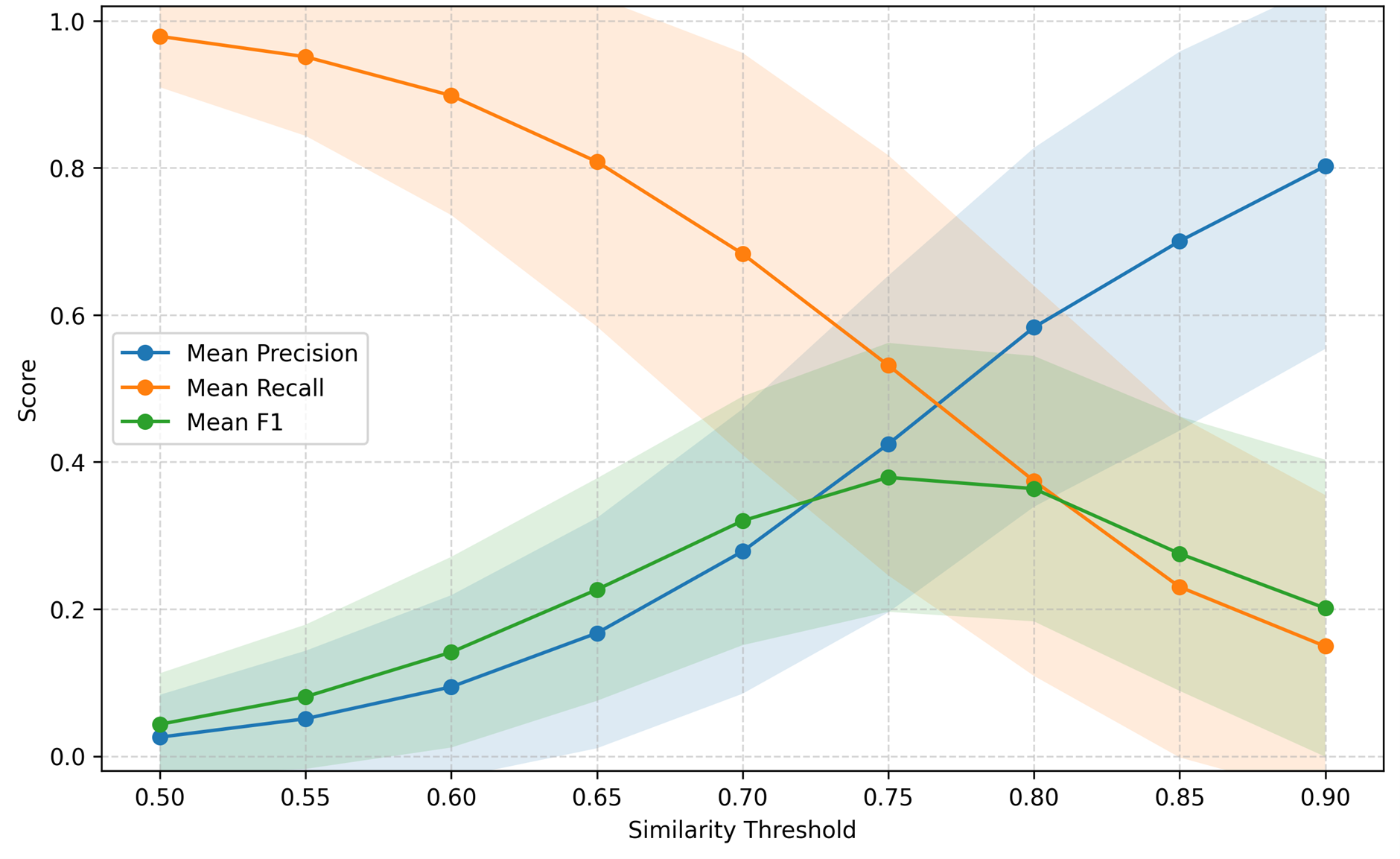}
  \caption{Mean $\pm$ SD: Precision, Recall, F1 versus similarity cut-off for Narrow Terms Retrieval (across SMQs).}
  \label{fig:narrow_perf}
\end{figure}

Although not explicitly optimized for narrow vs. broad PTs, narrow terms tended to yield higher similarity scores (see Figure~\ref{fig:narrow_perf}). Recall increased slightly in the narrow-term-only evaluation while precision declined modestly; the F1 curve shifted rightward by approximately +0.05, and the optimal F1 was observed at thresholds between 0.75 and 0.80.

\section{Discussion}
We developed and tested an automated MedDRA query (AMQ) system based on statistical similarity analysis of the SafeTerm MedDRA Map. The AMQ operates fully automatically and was not tuned or trained using SMQ or OCMQ reference sets. It relies purely on syntactic and semantic similarity to identify relevant Preferred Terms (PTs), making it objective, reproducible, and easily maintainable as MedDRA evolves. When evaluated against FDA OCMQs in a separate study, the system showed comparable performance patterns, confirming the robustness and generalisability of the underlying method across datasets.

The automatic similarity threshold selection performed close to the value that maximized the F1 score, but with slightly higher recall and lower precision. This supports a practical AMQ workflow: users can apply the auto-selected threshold to capture a broad set of candidate PTs, then review the retrieved terms in order of decreasing similarity for manual selection or refinement. This approach balances efficiency with accuracy, ensuring that both broad and narrow PTs are considered, while giving human reviewers the ability to prioritize clinically relevant terms. As expected, narrow, more specific PTs generally received higher similarity scores and ranked closer to the input query, guiding the prioritization process.

Broader clinical concepts, however, remain more challenging. Some SMQs describe sets of symptoms or complications rather than a clearly defined disorder, making it difficult for an algorithm to infer all possible downstream consequences. This illustrates a natural trade-off between specificity and coverage: expanding the scope improves sensitivity but can reduce precision. Importantly, these findings highlight that if a medical concept is already defined within MedDRA terminology, it should be expressed using the corresponding PTs rather than free-text labels when constructing SMQs or other standard queries. Doing so ensures clearer concept definition and enables more accurate and consistent retrieval by automated tools such as AMQ.

\section{Conclusions}
We present an automated MedDRA query (AMQ) system that combines AI-based embeddings with statistical similarity analysis. Validated against MedDRA Tier-1 SMQ v28.1, the system delivers balanced performance comparable to expert-defined query lists and provides a practical tool for pharmacovigilance and clinical safety applications. The method is fully unsupervised, reproducible, and adaptable to new MedDRA versions, requiring only the input query and the current dictionary.

For practical use, we recommend starting with the automated similarity threshold method to capture most relevant PTs (high recall), then reviewing the retrieved terms in order of decreasing similarity. Increasing the threshold can improve precision by prioritizing narrow, high-confidence PTs. Using well-defined MedDRA PT terminology for query concepts further enhances retrieval accuracy. Using a combination of PTs is also an option for SMQs that describe broader clinical manifestations. This approach provides an efficient, version-agnostic workflow that balances sensitivity and specificity while supporting expert review.

\bibliographystyle{unsrtnat}
\bibliography{main}

@article{brown1999meddra,
  title={The Medical Dictionary for Regulatory Activities (MedDRA)},
  author={Brown, E.G. and Wood, L. and Wood, S.},
  journal={Drug Safety},
  volume={20},
  number={2},
  pages={109--117},
  year={1999},
  doi={10.2165/00002018-199920020-00002}
}

@article{mozzicato2007smq,
  title={Standardised MedDRA Queries (SMQs): their role in signal detection},
  author={Mozzicato, P.},
  journal={Drug Safety},
  volume={30},
  number={7},
  pages={617--619},
  year={2007},
  doi={10.2165/00002018-200730070-00009}
}

@article{siegersma2022pipeline,
  title={Development of a Pipeline for Adverse Drug Reaction Identification in Clinical Notes: Word Embedding Models and String Matching},
  author={Siegersma, K.R. and Evers, M. and Bots, S.H. and Groepenhoff, F. and Appelman, Y. and Hofstra, L. and others},
  journal={JMIR Medical Informatics},
  volume={10},
  number={1},
  pages={e31063},
  year={2022},
  doi={10.2196/31063}
}

@article{lahiri2025benchmarking,
  title={Benchmarking Transformer Embedding Models for Biomedical Terminology Standardization},
  author={Lahiri, A. and Shukla, S. and Stear, B. and Mohseni Ahooyi, T. and Beigel, K. and Margolskee, E. and Taylor, D.},
  journal={Machine Learning with Applications},
  volume={21},
  pages={100683},
  year={2025},
  doi={10.1016/j.mlwa.2025.100683}
}

@article{vandenhende2025ocmq,
  title={Automated Generation of Custom MedDRA Queries Using SafeTerm Medical Map},
  author={Vandenhende, F. and Georgiou, M. and Psaras, T. and Georgiou, A. and Karekla, E. and Hadjicosta, E.},
  journal={Submitted manuscript},
  year={2025},
  note={Submitted for publication}
}

@article{proestel2025ocmq,
  title={The Development and Use of Office of New Drugs Custom Medical Queries for Safety Analyses of Clinical Trial Data},
  author={Proestel, S. and Popat, V. and Unger, E.F. and Jeng, L.J.B. and others},
  journal={Drug Safety},
  year={2025},
  doi={10.1007/s40264-025-01582-1}
}

@misc{fda2025ocmq,
  title={Office of New Drugs Custom Medical Queries (OCMQs)},
  author={{U.S. Food and Drug Administration}},
  year={2025},
  howpublished={\url{https://www.fda.gov/drugs/development-resources/office-new-drugs-custom-medical-queries-ocmqs}},
  note={[Accessed June 10, 2025]}
}

@misc{kneedle,
  title={Kneedle: An Algorithm for Detecting the Knee Point in Noisy Data},
  author={Satopaa, V. and Albrecht, J. and Irwin, D. and Raghavan, B.},
  year={2011},
  note={Algorithm referenced in methods (Knee method).}
}

\clearpage

\begin{longtable}{lcccccccccccc}
\caption{AMQ Performance Summary across 110 SMQs: Precision, Recall and F1 at multiple similarity cut-offs.}\label{tab:performance_summary} \\
\begin{tabular}{c|cccc|cccc|cccc}
\hline
\textbf{Similarity} & \multicolumn{4}{c|}{\textbf{Precision}} & \multicolumn{4}{c|}{\textbf{Recall}} & \multicolumn{4}{c}{\textbf{F1}} \\
\textbf{Cut-off} & Mean & SD & Min & Max & Mean & SD & Min & Max & Mean & SD & Min & Max \\
\hline
0.50 & 0.04 & 0.06 & 0.00 & 0.50 & 0.94 & 0.10 & 0.37 & 1.00 & 0.07 & 0.08 & 0.01 & 0.48 \\
0.55 & 0.08 & 0.10 & 0.01 & 0.75 & 0.86 & 0.17 & 0.24 & 1.00 & 0.13 & 0.11 & 0.02 & 0.65 \\
0.60 & 0.15 & 0.14 & 0.02 & 0.84 & 0.76 & 0.23 & 0.12 & 1.00 & 0.21 & 0.13 & 0.03 & 0.70 \\
0.65 & 0.25 & 0.17 & 0.01 & 0.88 & 0.62 & 0.27 & 0.04 & 1.00 & 0.30 & 0.14 & 0.03 & 0.66 \\
0.70 & 0.39 & 0.20 & 0.00 & 0.89 & 0.48 & 0.28 & 0.00 & 0.96 & 0.36 & 0.16 & 0.00 & 0.68 \\
0.75 & 0.56 & 0.21 & 0.00 & 1.00 & 0.34 & 0.24 & 0.00 & 0.94 & 0.35 & 0.18 & 0.00 & 0.78 \\
0.80 & 0.72 & 0.20 & 0.00 & 1.00 & 0.21 & 0.18 & 0.00 & 0.88 & 0.27 & 0.18 & 0.00 & 0.88 \\
0.85 & 0.81 & 0.19 & 0.00 & 1.00 & 0.11 & 0.12 & 0.00 & 0.72 & 0.17 & 0.15 & 0.00 & 0.80 \\
0.90 & 0.89 & 0.18 & 0.00 & 1.00 & 0.05 & 0.06 & 0.00 & 0.31 & 0.10 & 0.10 & 0.00 & 0.48 \\
\hline
\end{tabular}
\end{longtable}

\clearpage

\begin{table}[H]
\centering
\caption{Performance metrics (Mean \(\pm\) SD) comparing the automated similarity threshold selection with the threshold corresponding to maximum F1 across 110 SMQs.}
\label{tab:automated_vs_f1}
\centering
\begin{tabular}{lcc}
\hline
\textbf{Metric} & \textbf{Automated Similarity Method} & \textbf{At Threshold of Maximum F1} \\
\hline
Precision & $0.29 \pm 0.19$ & $0.45 \pm 0.18$ \\
Recall    & $0.58 \pm 0.26$ & $0.48 \pm 0.17$ \\
F1        & $0.32 \pm 0.15$ & $0.44 \pm 0.16$ \\
Threshold & $0.66 \pm 0.05$ & $0.70 \pm 0.07$ \\
\hline
\end{tabular}
\end{table}

\clearpage

\end{document}